\documentclass[11pt]{article}
\usepackage[margin=1in]{geometry}
\usepackage[T1]{fontenc}
\usepackage{lmodern}
\usepackage{microtype}
\usepackage{graphicx}
\usepackage{tikz}
\usetikzlibrary{arrows.meta,positioning,fit,calc,matrix}
\usepackage{booktabs}
\usepackage{longtable}
\usepackage{array}
\usepackage{amsmath}
\usepackage{url}
\usepackage{caption}
\usepackage{subcaption}
\usepackage{float}
\usepackage{enumitem}
\usepackage[numbers,sort&compress]{natbib}
\graphicspath{{figures/}}
\usepackage{makecell}
\usepackage[hidelinks]{hyperref}

\title{\textbf{Differential Proteomics to Symbolic Rules: An Audit of SHAP Guidance in Lung Adenocarcinoma}}
\author{Surajit Das\\ AI \& Mathematical Modeling\\ Independent Researcher, Kolkata, India\\
mr.surajitdas@gmail.com}
\date{}

\begin{document}
\maketitle

\begin{abstract}
We tested whether SHAP improves protein-based Horn rules for distinguishing
lung adenocarcinoma from adjacent-normal tissue. CPTAC provided 213 samples
(111 tumour, 102 adjacent-normal) from 111 patients, including 102 matched
pairs. Per-protein effect estimates and tests used the same complete pairs,
with Benjamini--Hochberg correction. Sparse logistic regression used five
stratified patient-grouped outer folds. Filtering, ranking, imputation,
scaling, feature selection and regularisation tuning were training-only.
Feature selection was repeated within inner folds. A separate five-fold
patient-grouped evaluation of first-order inductive logic programming
(FOL/ILP) used training-derived evidence and discretisation.

Of 9,518 proteins passing quality control, 378 differed in abundance
(40 up, 338 down) at FDR $<0.05$ and absolute mean paired log-ratio
difference $\geq1$. Classifier ROC-AUC, average precision, accuracy and
balanced accuracy were 0.9997, 0.9998, 0.9906 and 0.9906, respectively.
Held-out rule accuracy, balanced accuracy, precision, recall and F1 were
0.9765, 0.9767, 0.9818, 0.9730 and 0.9774, respectively. HSPA12B,
PALM2-AKAP2, STARD13 and PPM1F had non-zero coefficients in all five
classifier folds. On the original rule splits, one clause appeared twice and the others once.

On common folds across 111 patients,
F1 was 0.9600 with SHAP, 0.9910 without SHAP and 0.9909 for L1 logistic
regression. The inclusive SHAP cutoff admitted all 100 ranked proteins per
audited fold. Absolute coefficients reproduced candidate sets, clauses and
held-out predictions in both analyses. A single threshold reproduced
full-rule predictions in the paired-only cohort.

Within-cohort discrimination leaves diagnostic utility, early detection,
prognosis and external generalisability unestablished. Biomarker and clinical
claims require independent validation, rule-stability testing under repeated
resampling and deeper biological annotation.
\end{abstract}


\section{Introduction}
Lung adenocarcinoma (LUAD), a major histological subtype of lung cancer,
has been profiled proteogenomically alongside matched adjacent-normal
tissue, providing a resource for studying tumour biology and protein
alterations linked to molecular and cellular processes
\citep{gillette2020,pdc2026}. These data support methodological development.
Clinical screening and diagnosis require evaluation in independent
populations.

Machine learning can prioritise proteins for classification, and SHAP
provides additive explanations of individual predictions
\citep{lundberg2017,mann2021,sun2026}. In high-dimensional biomedical data,
biomarker discovery and validation require careful feature selection and
resampling \citep{diaz2022,lewis2023,parvandeh2020,lustgarten2009}.
Inductive logic programming (ILP) expresses combinations of evidence as
symbolic rules \citep{muggleton1991,zhang2024}.

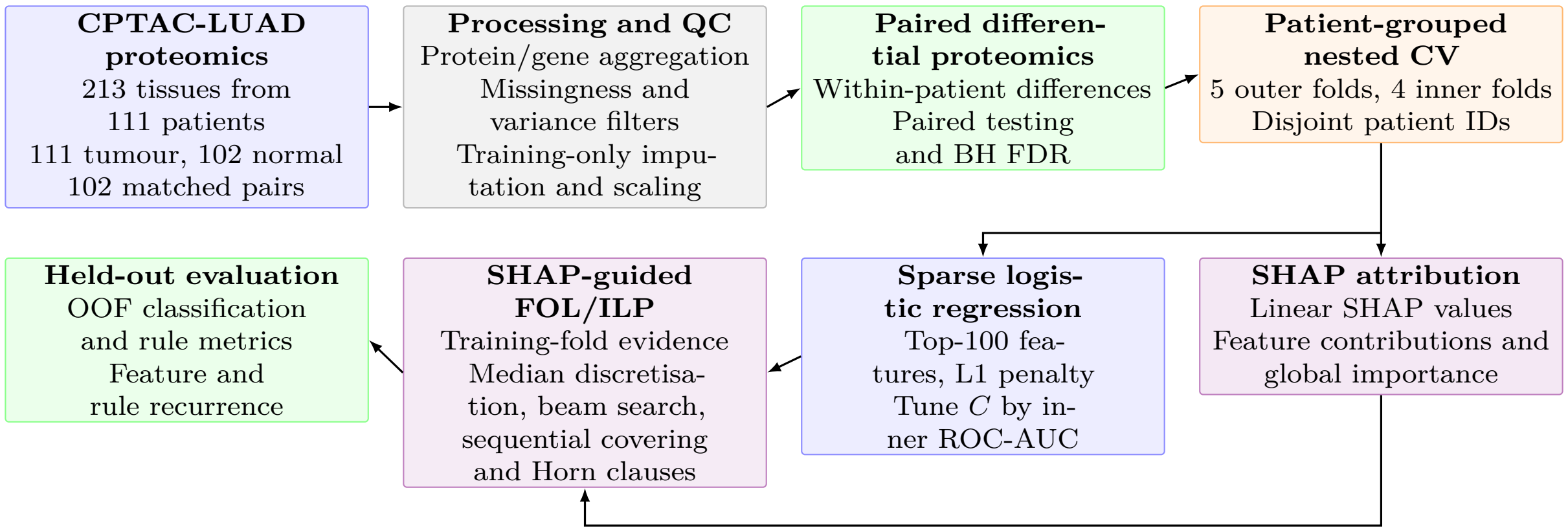
\begin{figure}[t]
\centering
\begin{tikzpicture}[
  block/.style={
    draw,
    rounded corners=1pt,
    align=center,
    text width=.216\linewidth,
    inner xsep=2pt,
    inner ysep=2pt,
    font=\scriptsize
  },
  flow/.style={-{Latex[length=1.4mm]}, semithick}
]
\matrix (workflow) [
  matrix of nodes,
  nodes in empty cells,
  row sep=5mm,
  column sep=.021\linewidth
] {
  |[block, fill=blue!7, draw=blue!45]| {
    \textbf{CPTAC-LUAD proteomics}\\
    213 tissues from 111 patients\\
    111 tumour, 102 normal\\
    102 matched pairs
  }
  &
  |[block, fill=gray!10, draw=gray!55]| {
    \textbf{Processing and QC}\\
    Protein/gene aggregation\\
    Missingness and variance filters\\
    Training-only imputation and scaling
  }
  &
  |[block, fill=green!8, draw=green!45]| {
    \textbf{Paired differential proteomics}\\
    Within-patient differences\\
    Paired testing and BH FDR
  }
  &
  |[block, fill=orange!8, draw=orange!55]| {
    \textbf{Patient-grouped nested CV}\\
    5 outer folds, 4 inner folds\\
    Disjoint patient IDs
  }
  \\
  |[block, fill=green!8, draw=green!45]| {
    \textbf{Held-out evaluation}\\
    OOF classification and rule metrics\\
    Feature and rule recurrence
  }
  &
  |[block, fill=violet!8, draw=violet!50]| {
    \textbf{SHAP-guided FOL/ILP}\\
    Training-fold evidence\\
    Median discretisation, beam search,\\
    sequential covering and Horn clauses
  }
  &
  |[block, fill=blue!7, draw=blue!45]| {
    \textbf{Sparse logistic regression}\\
    Top-100 features, L1 penalty\\
    Tune $C$ by inner ROC-AUC
  }
  &
  |[block, fill=violet!9, draw=violet!50]| {
    \textbf{SHAP attribution}\\
    Linear SHAP values\\
    Feature contributions and\\
    global importance
  }
  \\
};

\draw[flow] (workflow-1-1.east) -- (workflow-1-2.west);
\draw[flow] (workflow-1-2.east) -- (workflow-1-3.west);
\draw[flow] (workflow-1-3.east) -- (workflow-1-4.west);

\node[
  fit=(workflow-1-1)(workflow-1-2)(workflow-1-3)(workflow-1-4),
  inner sep=0pt, outer sep=0pt
] (upperrow) {};
\node[
  fit=(workflow-2-1)(workflow-2-2)(workflow-2-3)(workflow-2-4),
  inner sep=0pt, outer sep=0pt
] (lowerrow) {};

\coordinate (rowgap) at ($(upperrow.south)!0.5!(lowerrow.north)$);
\coordinate (cvbranch) at (workflow-1-4.south |- rowgap);
\draw[flow] (workflow-1-4.south) -- (workflow-2-4.north);
\draw[flow] (cvbranch) -| (workflow-2-3.north);

\draw[flow] (workflow-2-3.west) -- (workflow-2-2.east);
\coordinate (returnlevel) at ([yshift=-2.5mm]workflow.south);
\draw[flow] (workflow-2-4.south)
  -- (workflow-2-4.south |- returnlevel)
  -| (workflow-2-2.south);

\draw[flow] (workflow-2-2.west) -- (workflow-2-1.east);
\end{tikzpicture}

\caption{CPTAC-LUAD workflow from proteomic quality control and paired
differential analysis through patient-grouped nested classification, SHAP
attribution, and SHAP-guided first-order logic/inductive logic programming.
Full-cohort differential analysis is descriptive. Within cross-validation,
predictive feature selection, preprocessing, attribution evidence, and rule
induction are confined to training partitions. Held-out folds are reserved
for evaluation.}
\label{fig:workflow}
\end{figure}

\begin{figure}[t]
\centering
\begin{tikzpicture}[
  node distance=4mm,
  stage/.style={
    draw=black!55,
    rounded corners=1pt,
    fill=black!4,
    align=center,
    font=\scriptsize,
    inner xsep=3pt,
    inner ysep=2pt
  },
  flow/.style={-{Latex[length=1.5mm]}, semithick}
]
\node[stage] (diff) {\textbf{Paired proteomics}\\Direction and magnitude};
\node[stage, right=of diff] (shap)
  {\textbf{Sparse model + SHAP}\\Protein contributions};
\node[stage, right=of shap] (rules)
  {\textbf{FOL/ILP}\\Auditable Horn clauses};

\draw[flow] (diff.east) -- (shap.west);
\draw[flow] (shap.east) -- (rules.west);
\end{tikzpicture}
\caption{Three-stage explainability chain. Patient-matched differential proteomics estimates the direction and magnitude of molecular changes. Sparse modelling and SHAP quantify proteins’ contributions to model predictions. FOL/ILP converts training-derived evidence into auditable Horn clauses. Patient-grouped outer folds keep training and evaluation patients disjoint.}
\label{fig:explainabilitychain}
\end{figure}
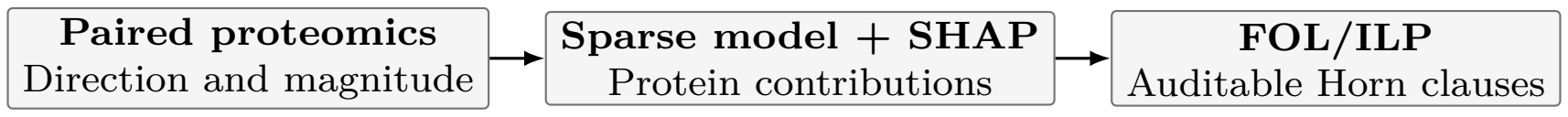

Our three-stage explainability workflow (Figures~\ref{fig:workflow}
and~\ref{fig:explainabilitychain}) uses patient-matched differential
proteomics to identify protein abundance changes and their direction.
Sparse logistic regression identifies proteins associated with tissue
discrimination, and ranked linear-model SHAP attributions quantify the
magnitude and direction of each protein's contribution to individual
predictions. First-order logic and ILP (FOL/ILP) combine paired differential
evidence and SHAP attributions from training folds into compact, inspectable
relational Horn clauses. The stages address distinct questions about
tissue-level abundance, predictive importance and how evidence combines
in rules. Causal effects, biological mechanisms and LUAD biomarkers
require separate validation. We separate full-cohort descriptive summaries
from evaluation on held-out patients and report feature and clause
recurrence across five outer folds to assess stability.

We used matched patient-grouped folds to test whether linear-model SHAP
changes the proteins selected, the Horn clauses learned, or the resulting
held-out predictions. Absolute L1 coefficients reproduced all three in
both CPTAC-LUAD analyses, and a single-protein threshold reproduced the
full rule learner's predictions in the paired-only cohort. The workflow
records the measured abundance and training-derived evidence behind
individual rule decisions, allowing those decisions to be replayed.

\section{Methodology}

\textbf{\textit{\underline{Cohort and proteomic data:}}}
Sample annotations and the biospecimen manifest in the supplied CPTAC-LUAD
files identified 213 tissues from 111 patients (111 primary tumours and
102 adjacent solid-tissue normals, forming 102 matched pairs).
Classification distinguished these tissues within the cohort. Pairing
focuses the descriptive contrast on within-patient differences and
partially controls inherited background, exposure history, and other
between-patient variation. Tissue composition, tumour purity, batch effects,
and clinical confounding can persist. Adjacent-normal tissue differs from
tissue from cancer-free individuals. TMT log-ratios were collapsed by gene
median and aggregated by aliquot as specified in the notebook.

\textbf{\textit{\underline{Paired differential proteomics:}}}
After averaging replicates within patient and tissue type, tissue means
and tests for each protein $p$ used the same patients $i$ with finite
measurements in both tissues,
$d_{ip}=\bar{x}_{i,p}^{\mathrm{tumour}}-\bar{x}_{i,p}^{\mathrm{normal}}$.
A one-sample $t$-test on $d_{ip}$ required at least 10 complete pairs.
Benjamini--Hochberg correction covered all tested proteins
\citep{benjamini1995}. Descriptive differential abundance required FDR
$<0.05$ and an absolute mean paired difference $\geq1$. Positive and
negative differences indicated increased and decreased tumour abundance,
respectively. Full-cohort results were descriptive. Both evaluations
used their corresponding training partitions for feature ranking and
evidence construction.

\textbf{\textit{\underline{Nested sparse classification:}}}
Protein measurements $\mathbf{x}_i$ predicted tumour ($y_i=1$) or
adjacent-normal tissue ($y_i=0$), with
$p_i=P(y_i=1\mid\mathbf{x}_i)=
[1+\exp(-\beta_0-\boldsymbol{\beta}^{T}\mathbf{x}_i)]^{-1}$
and loss
$\mathcal L=-\sum_{i=1}^{n}w_{y_i}
[y_i\log p_i+(1-y_i)\log(1-p_i)]+\lambda\|\boldsymbol{\beta}\|_1$.
The L1 penalty encourages zero coefficients and reduces model complexity.
Scikit-learn's \texttt{LogisticRegression} used \texttt{penalty='l1'},
\texttt{liblinear}, balanced class weights $w_{y_i}$, up to 5,000 iterations,
and seed 42. We tested $C\in\{0.01,0.03,0.1,0.3,1,3,10\}$, with larger
values weakening regularisation.

Stratified patient-grouped cross-validation kept patients disjoint at
each split (five outer and four inner folds). Each inner training
partition independently fitted quality filtering ($\leq20\%$ missingness,
non-zero variance), paired differential ranking, top-100 selection,
median imputation, and standardisation for its validation partition.
Mean inner ROC-AUC selected $C$. Feature selection and preprocessing were
refitted on the full outer training partition before held-out evaluation.
Pooled out-of-fold probabilities yielded ROC-AUC and average precision.
Labels at 0.5 yielded accuracy, balanced accuracy, precision, sensitivity,
specificity, and F1.

\textbf{\textit{\underline{SHAP and symbolic rule induction:}}}
SHapley Additive exPlanations (SHAP) used an independent training-background
mean. For standardised abundance $z_{ij}$ and fitted coefficient $\beta_j$,
$\phi_{ij}=\beta_j(z_{ij}-\bar z_{j,\mathrm{background}})$ and
$I_j=n^{-1}\sum_{i=1}^{n}|\phi_{ij}|$.
The full-cohort importance plot ranks $I_j$, and the beeswarm shows the
distribution of signed contributions, one dot per tissue sample. Both
describe fitted model behaviour. Paired effects, causal estimates, and
independent validation require separate analyses.

A separate five-fold stratified patient-grouped FOL/ILP evaluation confined
quality filtering, paired evidence, nested tuning, fitting, SHAP, and
knowledge-base construction to outer training data. Differential and
attribution evidence selected up to 14 proteins. Training medians imputed
missing values before median discretisation. Held-out samples used the same
imputation values and thresholds. The knowledge base encoded abundance
states and protein evidence (Table~\ref{tab:predicates}). Each Horn clause
$\mathrm{tumor}(X)\leftarrow L_1(X),\ldots,L_k(X)$ predicted tumour when all
its literals, representing abundance states or evidence conditions, held.

\begin{table}[t]
\centering
\caption{Principal FOL predicates used for rule induction.}
\label{tab:predicates}
\small
\begin{tabular}{p{0.25\linewidth}p{0.65\linewidth}}
\toprule
Predicate & Operational meaning\\
\midrule
$\mathrm{high}(X,P)$ / $\mathrm{low}(X,P)$ &
Abundance of protein $P$ in tissue sample $X$ is above / at or below its
training median.\\
$\mathrm{up\_in\_tumor}(P)$ / $\mathrm{down\_in\_tumor}(P)$ &
Training-fold paired mean difference for $P$ is positive / negative.\\
$\mathrm{differential}(P)$ & Training-fold FDR for $P$ is $<0.05$.\\
$\mathrm{high\_shap}(P)$ &
Mean absolute training SHAP for $P$ reaches the prespecified 60th percentile.\\
$\mathrm{stable}(P)$ & $P$ has a non-zero L1 coefficient in the corresponding
training-fold model.\\
\bottomrule
\end{tabular}
\end{table}

For example,
$\mathrm{tumor}(X)\leftarrow\mathrm{low}(X,P),
\mathrm{down\_in\_tumor}(P),\mathrm{high\_shap}(P)$
classifies $X$ as tumour if at least one protein $P$ satisfies all three
predicates. The clause has no causal interpretation.

Beam search (width 30) allowed up to three terms per clause, counting a
relational pattern sharing a protein variable as one term. Eligible clauses
covered at least four positive training examples with precision $\geq0.75$.
The score was $S(R)=F_1(R)-0.015\max(0,|R|-1)$, where $|R|$ counts terms.
Sequential covering removed covered positives after each selection and
stopped at four clauses or when no eligible clause or sufficient positive
coverage remained. Rules and thresholds were applied unchanged to held-out
folds. Pooled predictions yielded accuracy, balanced accuracy, precision,
recall, and F1. Induction precision and recall were training diagnostics.

\textbf{\textit{\underline{Stability and implementation:}}}
Across five outer folds, we tabulated descriptive recurrence measures:
counts and proportions for top-100 selection and non-zero L1 coefficients,
exact-clause counts with mean training precision and recall, and
candidate-protein recurrence. We used NumPy, pandas, SciPy, scikit-learn,
statsmodels, and Matplotlib. The base seed was 42, with seed 59 for
rule-learner outer splitting and inner tuning. Evaluation was internal to
CPTAC-LUAD.

\begin{figure}[H]
\centering
\includegraphics[width=0.46\linewidth]{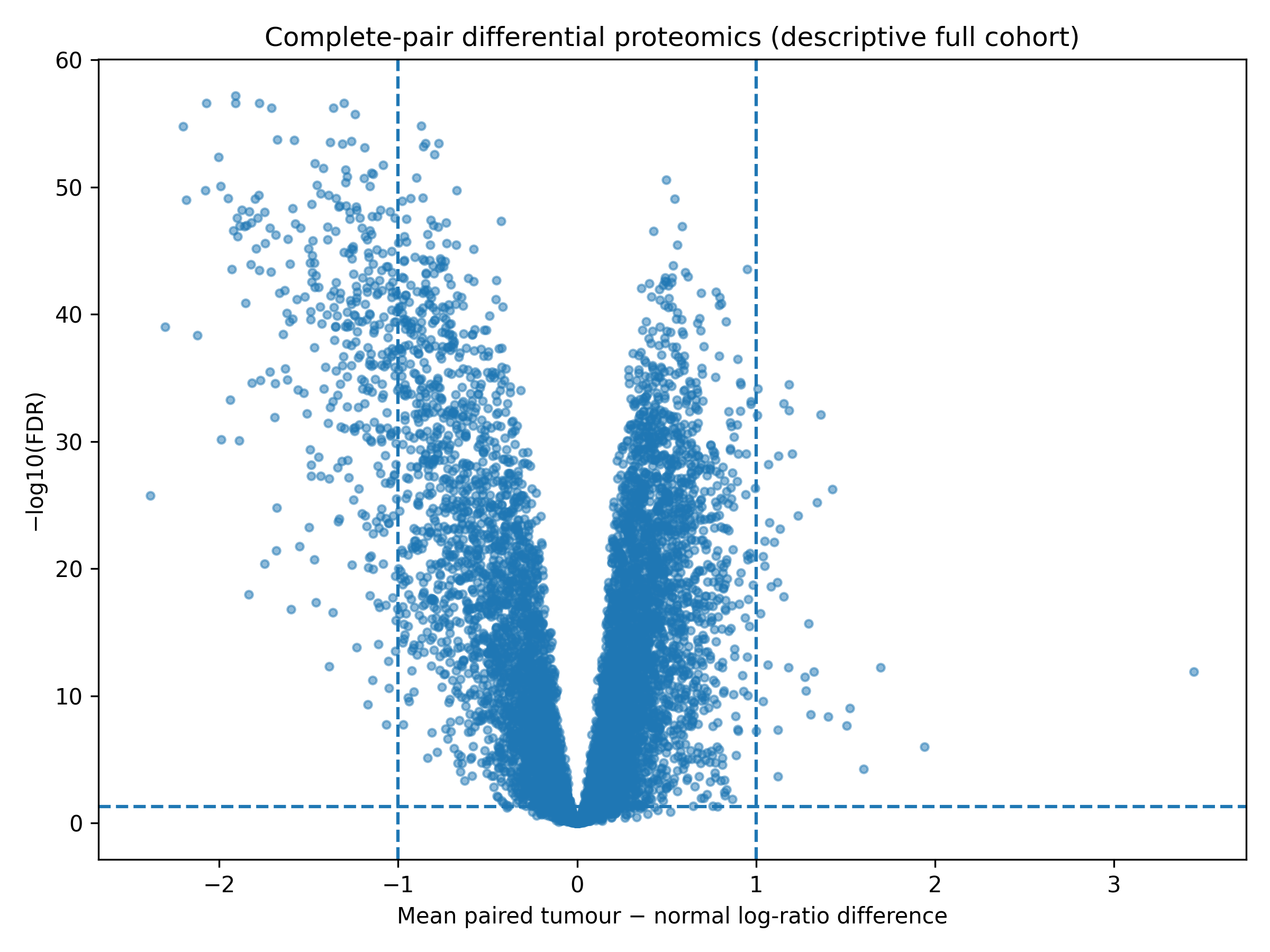}
\caption{Descriptive full-cohort paired differential proteomics using
protein-specific complete pairs. Held-out validation was separate.}
\label{fig:volcano}
\end{figure}

\textbf{\textit{\underline{Matched comparators and uncertainty:}}}
We retained the original separate splits to document the submitted model
and rule results. Direct comparisons used scikit-learn 1.8.0 and seed 59
for outer and inner grouping. L1 tuning maximised mean inner ROC-AUC over
the seven-value grid, resolving exact ties in favour of the smallest $C$.
All arms shared five patient-grouped outer folds within each cohort.
Sensitivity analysis repeated the procedure for the 102 matched patients.

The full learner retained the first 14 of the top 100 proteins ordered by
the inclusive 60th-percentile predicate, descending mean absolute SHAP,
then ascending differential FDR. The absolute-coefficient comparator
replaced $I_j$ with $|\beta_j|$ in the predicate and ranking while keeping
the clause search identical. The no-SHAP arm disabled the attribution
predicate, retained the non-zero-coefficient predicate, and selected 14
proteins by paired differential rank. The arm without differential
predicates retained candidate selection and restricted search to
fixed-protein terms. The single-threshold baseline used the full learner's
candidates and medians to select the eligible fixed-protein term with
highest training F1. A depth-three tree (at least four samples per leaf)
and a 300-tree random forest used the L1 model's preprocessed top-100
features, balanced class weights, and seed 42. Probabilistic models used
a 0.5 decision threshold. A channel-only control learned majority tissue
labels by TMT channel on the same training folds, using the training
class majority for unseen channels and majority voting across distinct
channels for replicated aliquots (ties favoured tumour).

For all methods, pooled out-of-fold tissue predictions within each cohort
yielded balanced accuracy and F1. Percentile intervals for F1 differences
used 1,000 patient-level bootstrap draws of held-out predictions (seed 42),
keeping paired tissues together. These intervals condition on the fitted
cross-validation predictions and exclude refitting variability. The
CPTAC-LUAD benchmark compared internally implemented baselines and
ablations. Published SHAP and ILP systems served as methodological
context. For the rule learner, we counted fixed-protein and relational
clauses and traced a held-out decision to its threshold and measured
abundance.

\section{Results}

\textbf{\textit{\underline{Cohort and paired differential abundance:}}}
The cohort comprised 213 tissues from 111 patients (111 tumour,
102 adjacent-normal, 102 matched pairs). Full-cohort filtering retained
9,518 proteins. Of these, 378 met the prespecified descriptive criteria
(FDR $<0.05$, absolute mean paired difference $\geq1$), with 40 increased
and 338 decreased in tumour. Protein-specific missingness left a median
of 102 complete pairs (range 79--102, Figure~\ref{fig:volcano}).

\textbf{\textit{\underline{Nested patient-grouped sparse classification:}}}
Pooled outer-fold predictions gave ROC-AUC 0.9997, average precision
0.9998, accuracy and balanced accuracy 0.9906, precision, sensitivity,
and F1 0.9910, and specificity 0.9902 (Table~\ref{tab:ml}). These
near-ceiling scores describe controlled tumour-versus-adjacent-normal
tissue classification within one cohort. Performance in blood,
early-stage screening, independent clinical populations, and clinical
decision pathways remains untested.

\begin{table}[H]
\centering
\begin{minipage}[t]{0.48\linewidth}
\centering
\caption{Pooled out-of-fold sparse logistic performance on the original splits.}
\label{tab:ml}
{\small
\setlength{\tabcolsep}{3pt}
\begin{tabular}{@{}lr@{\hspace{0.8em}}lr@{}}
\toprule
Metric & Value & Metric & Value\\
\midrule
ROC-AUC            & 0.9997 & Precision   & 0.9910\\
Average precision  & 0.9998 & Sensitivity & 0.9910\\
Accuracy           & 0.9906 & Specificity & 0.9902\\
Balanced accuracy  & 0.9906 & F1          & 0.9910\\
\bottomrule
\end{tabular}%
}
\end{minipage}
\hfill
\begin{minipage}[t]{0.48\linewidth}
\centering
\caption{Pooled held-out FOL/ILP performance on the original splits.}
\label{tab:ilp}
\small
\setlength{\tabcolsep}{2pt}
\begin{tabular*}{\linewidth}{@{\extracolsep{\fill}}lrlr@{}}
\toprule
Metric & Value & Metric & Value\\
\midrule
Accuracy          & 0.9765 & F1 & 0.9774\\
Balanced accuracy & 0.9767 &    &       \\
Precision         & 0.9818 &    &       \\
Recall            & 0.9730 &    &       \\
\bottomrule
\end{tabular*}
\end{minipage}
\end{table}

\begin{figure}[H]
\centering
\begin{minipage}[c]{0.46\linewidth}
\centering
\includegraphics[width=\linewidth]{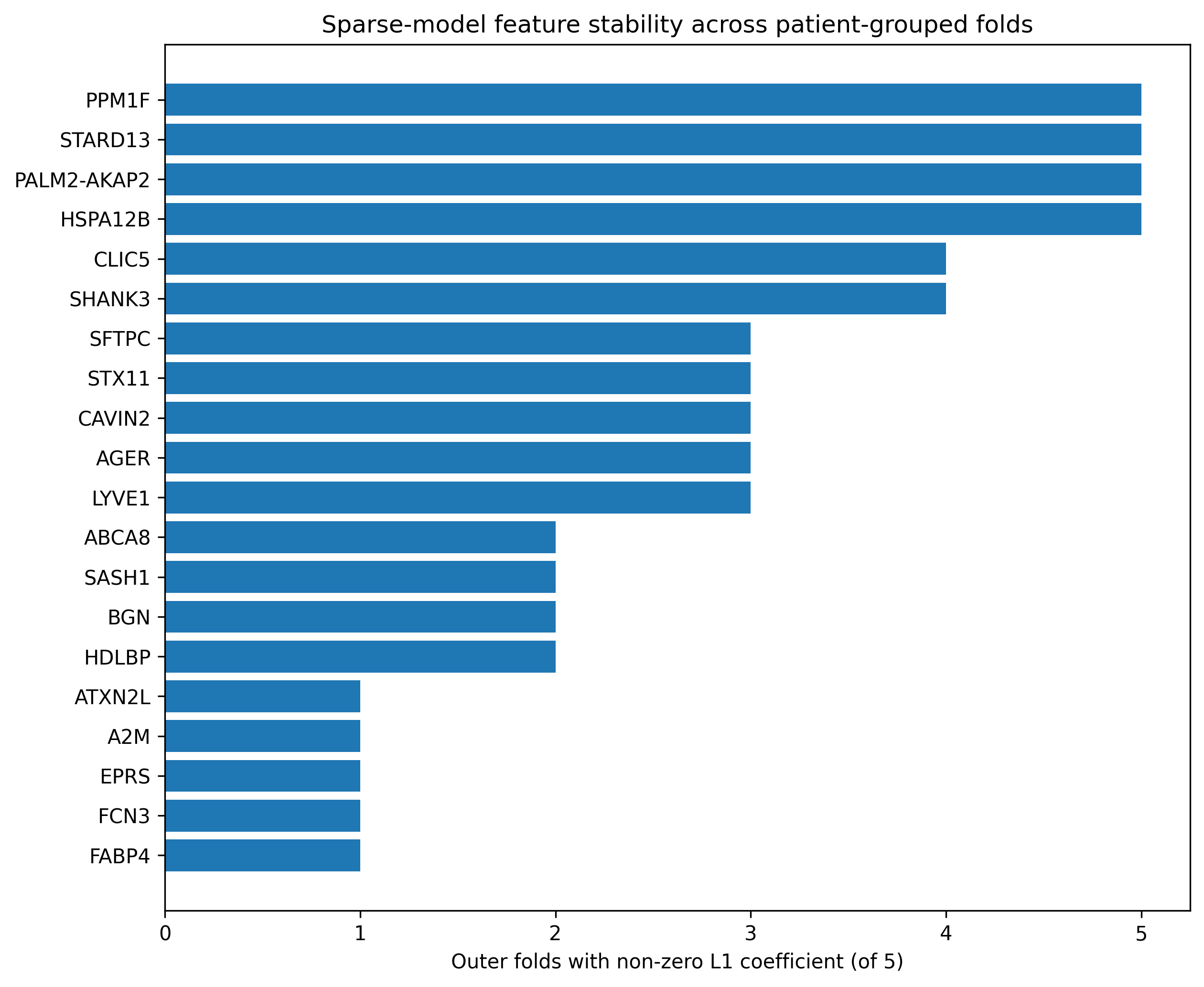}
\caption{Sparse-model feature recurrence across five patient-grouped outer folds.}
\label{fig:featurestability}
\end{minipage}%
\hspace{2mm}%
\begin{minipage}[c]{0.46\linewidth}
\centering
\captionof{table}{Top-100 selection and non-zero L1 coefficients across outer folds.}
\label{tab:featurestability}
\small
\begin{tabular}{lcc}
\toprule
Protein & \makecell{Top-100\\folds} & \makecell{Non-zero\\L1 folds}\\
\midrule
HSPA12B & 5/5 & 5/5\\
PALM2-AKAP2 & 5/5 & 5/5\\
STARD13 & 5/5 & 5/5\\
PPM1F & 5/5 & 5/5\\
SHANK3 & 5/5 & 4/5\\
CLIC5 & 5/5 & 4/5\\
AGER & 5/5 & 3/5\\
SFTPC & 5/5 & 3/5\\
\bottomrule
\end{tabular}
\end{minipage}
\end{figure}

The final descriptive signature included the eight proteins in
Table~\ref{tab:featurestability} among its non-zero coefficients. All
eight ranked in the top 100 in every outer fold. HSPA12B, PALM2-AKAP2,
STARD13, and PPM1F had non-zero coefficients in all five folds, compared
with four for SHANK3 and CLIC5 and three for AGER and SFTPC. The first
four recurred more often than any exact clause (maximum 2/5 folds).
These frequencies describe recurrence under the five-fold design.
Independent replication remains untested (Figure~\ref{fig:featurestability}).

\begin{figure}[H]
\centering
\caption{Exploratory linear-model SHAP attributions for prioritised proteins
from the descriptive full-cohort sparse logistic model. These plots use
the fitted cohort. Held-out and external validation require separate data.}
\label{fig:shap}
\begin{subfigure}[b]{0.42\linewidth}
\centering
\includegraphics[width=\linewidth]{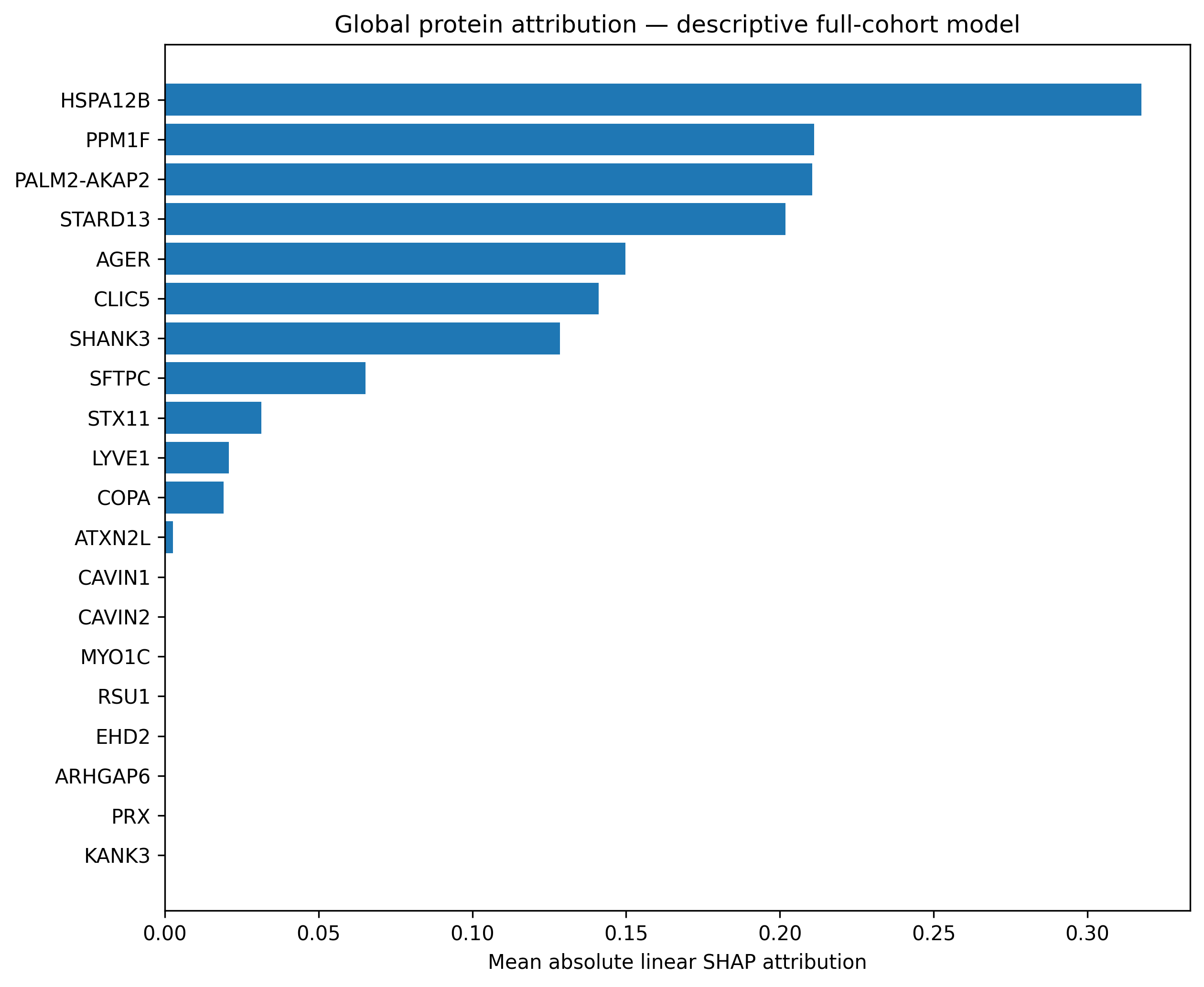}
\caption{Ranking by global mean absolute attribution.}
\label{fig:shapbar}
\end{subfigure}
\hfill
\begin{subfigure}[b]{0.48\linewidth}
\centering
\includegraphics[width=\linewidth]{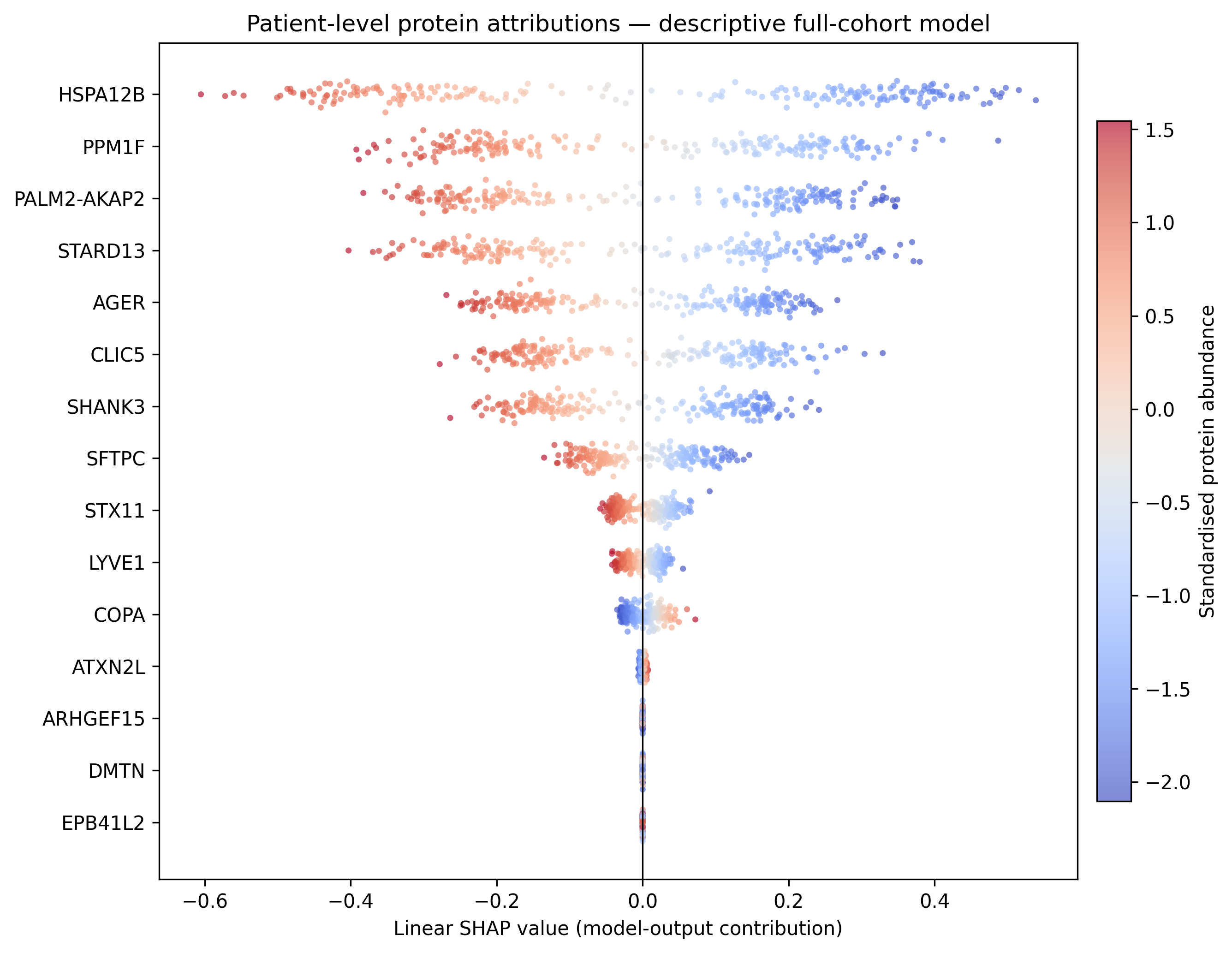}
\caption{Distribution and direction of sample-level attributions.}
\label{fig:shapbeeswarm}
\end{subfigure}
\end{figure}

Figure~\ref{fig:shap} shows the global ranking and sample-level distribution of SHAP attributions for prioritised proteins in the descriptive full-cohort model.

\textbf{\textit{\underline{Biological interpretation of the learned symbolic rules:}}}
Rules combine three evidence types. $\mathrm{high}(X,P)$ and
$\mathrm{low}(X,P)$ place protein $P$ in sample $X$ above or at/below its
training-derived median. $\mathrm{up\_in\_tumor}(P)$ and
$\mathrm{down\_in\_tumor}(P)$ encode the sign of the training paired
tumour--normal mean difference. $\mathrm{high\_shap}(P)$ uses the 60th
percentile of training-fold mean absolute SHAP. Together, these predicates
link sample abundance, cohort-derived direction, and model attribution.
Causality and biological mechanisms remain unestablished.

The descriptive full-cohort clause
$\mathrm{tumor}(X)\leftarrow\mathrm{high}(X,P)\land
\mathrm{up\_in\_tumor}(P)\land\mathrm{high\_shap}(P)$
fired for all 111 tumour tissues and one normal tissue (111 true positives,
one false positive, precision $111/112=0.991$). Both the rule and its
evidence came from the full cohort, so this is an in-sample result.
The most recurrent fold-derived clause combined low abundance,
down-in-tumour evidence, and high SHAP, appearing in two of five folds
(Table~\ref{tab:rulestability}). Each outer-training partition supplied
its own evidence, thresholds, and clauses for its held-out patients.
This accounts for the difference from the full-cohort summary. Exact-clause
recurrence measures repeatability under this design. Biological recurrence
and external validity remain untested.

\textbf{\textit{\underline{Held-out FOL/ILP performance and rule stability:}}}
On its original, separate outer splits, FOL/ILP achieved pooled held-out
accuracy 0.9765, balanced accuracy 0.9767, precision 0.9818, recall 0.9730,
and F1 0.9774 (Table~\ref{tab:ilp}). Table~\ref{tab:rulestability} lists
all four distinct clauses. The matched benchmark below uses common splits.

\begin{table}[H]
\centering
\caption{Exact Horn-clause recurrence across five outer folds.
Only clause bodies are shown. The head is $\mathrm{tumor}(X)$.}
\label{tab:rulestability}
\begin{tabular}{@{}p{0.64\linewidth}cc@{}}
\toprule
Clause body & Folds & Frequency\\
\midrule
\texttt{low(X,P), down\_in\_tumor(P), high\_shap(P)} & 2/5 & 0.40\\
\texttt{low(X,hspa12b)} & 1/5 & 0.20\\
\texttt{low(X,stard13)} & 1/5 & 0.20\\
\texttt{low(X,P), down\_in\_tumor(P), stable(P)} & 1/5 & 0.20\\
\bottomrule
\end{tabular}
\end{table}

\textbf{\textit{\underline{Matched classification and ablations:}}}
The scikit-learn 1.8.0 rerun used common outer splits, distinct from
the original evaluations.
Table~\ref{tab:matched} summarises pooled held-out performance across models and rule variants on the common patient-grouped folds. The full-minus-no-SHAP F1 difference was
$-0.0310$ (95\% patient-bootstrap interval $-0.0533$ to $-0.0090$)
across all patients. In the paired-only analysis, the full learner
and single-protein threshold made identical predictions. TMT channel
alone reached F1 0.9911 overall and 0.9903 in paired patients.

\begin{table}[H]
\centering\small
\caption{Pooled held-out performance on common patient-grouped folds.
BA denotes balanced accuracy. All-patient L1 ROC-AUC was 0.9999.}
\label{tab:matched}
\begin{tabular}{lrrrr}
\toprule
& \multicolumn{2}{c}{All 111 patients} & \multicolumn{2}{c}{102 paired patients}\\
Method & BA & F1 & BA & F1\\
\midrule
Random forest & 0.9902 & 0.9911 & 0.9853 & 0.9855\\
L1 logistic regression & 0.9910 & 0.9909 & 0.9853 & 0.9852\\
Depth-three tree & 0.9530 & 0.9550 & 0.9755 & 0.9754\\
Single-protein threshold & 0.9685 & 0.9674 & 0.9804 & 0.9802\\
Without differential predicates & 0.9636 & 0.9630 & 0.9804 & 0.9802\\
Without SHAP & 0.9906 & 0.9910 & 0.9755 & 0.9751\\
Absolute-coefficient rules & 0.9571 & 0.9600 & 0.9804 & 0.9802\\
Full SHAP-guided rules & 0.9571 & 0.9600 & 0.9804 & 0.9802\\
\bottomrule
\end{tabular}
\end{table}

\textbf{\textit{\underline{Attribution and regularisation audit:}}}
Linear SHAP gives
$I_j=|\beta_j|\,\overline{|z_j-\bar z_{j,\mathrm{background}}|}$.
Standardisation fixes variance, while mean absolute deviation can vary
by protein, so SHAP and coefficient rankings can differ. The inclusive
cutoff was zero in every fold of both analyses. All 100 proteins passed
in each all-patient fold, including those with zero coefficients. Ranking
still selected 14. Absolute coefficients reproduced the candidate sets,
clauses, and predictions in both analyses (Table~\ref{tab:gate} and
supplementary CSVs). Table~\ref{tab:tune} gives all primary inner scores.
The supplementary tuning CSV gives full-precision and paired-only results.

\begin{table}[H]
\centering\small
\caption{Attribution cutoff and counts across five all-patient folds.
Each fold starts with 100 proteins. Overlap counts candidates shared
with the absolute-coefficient comparator.}
\label{tab:gate}
\begin{tabular}{rrrrrr}
\toprule
Fold & Nonzero $\beta$ & Cutoff & Passing & Passing in top 14 & Overlap\\
\midrule
1 & 9 & 0 & 100 & 14 & 14\\
2 & 10 & 0 & 100 & 14 & 14\\
3 & 17 & 0 & 100 & 14 & 14\\
4 & 9 & 0 & 100 & 14 & 14\\
5 & 9 & 0 & 100 & 14 & 14\\
\bottomrule
\end{tabular}
\end{table}

\begin{table}[H]
\centering\small
\caption{Mean inner ROC-AUC by $C$ for all patients. Bold marks selection.}
\label{tab:tune}
\begin{tabular}{rrrrrrrr}
\toprule
Fold & 0.01 & 0.03 & 0.1 & 0.3 & 1 & 3 & 10\\
\midrule
1 & 0.500000 & \textbf{1.000000} & 1.000000 & 1.000000 & 1.000000 & 1.000000 & 1.000000\\
2 & 0.500000 & \textbf{1.000000} & 1.000000 & 1.000000 & 1.000000 & 1.000000 & 1.000000\\
3 & 0.500000 & 0.999432 & \textbf{1.000000} & 1.000000 & 1.000000 & 1.000000 & 1.000000\\
4 & 0.500000 & \textbf{1.000000} & 1.000000 & 1.000000 & 1.000000 & 1.000000 & 1.000000\\
5 & 0.500000 & \textbf{1.000000} & 1.000000 & 1.000000 & 1.000000 & 1.000000 & 1.000000\\
\bottomrule
\end{tabular}
\end{table}

\textbf{\textit{\underline{Rule complexity and held-out trace:}}}
In the matched rerun, each all-patient fold produced one clause with one
search term. Two used fixed proteins (HSPA12B and SHANK3), and three
used an existential protein variable, allowing one term to match multiple
eligible proteins. Two folds used low abundance, negative paired
direction, and a non-zero coefficient. One used the SHAP predicate.
All five paired-only folds yielded fixed-protein thresholds.

In all-patient fold 2, the HSPA12B clause fired at abundance
$\leq -0.3656$, the training-derived threshold. One held-out tumour
had measured abundance $-1.0913$ without imputation, satisfying the
low-state predicate and yielding a tumour prediction. The supplement
provides the fired-rule witnesses used to reconstruct this decision.

\section{Discussion}
Patient grouping and training-only preprocessing and selection support
internal evaluation of the paired-proteomics, linear SHAP, and ILP
workflow \citep{lundberg2017,muggleton1991}. Training-derived evidence
makes each clause auditable. Full-cohort differential results, SHAP
plots, and the high-abundance clause are descriptive. Four proteins
retained non-zero coefficients across all five classifier folds, while
the most recurrent original clause appeared in two rule folds.
Repeated grouped resampling and external evaluation are needed to
assess stability.

The 102 matched pairs reduce between-patient variation, with protein-specific pair counts varying because of missingness. Nine tumour-only patients contribute to classification alone. The paired-only analysis provides a sensitivity check for excluding those nine patients. Tumour purity, tissue composition, stage, and batch can affect abundance. Channel alone nearly matched the proteomic classifiers, so high internal scores cannot establish discrimination independent of channel. Screening asymptomatic individuals, early detection, malignant versus benign discrimination, plasma markers, prognosis, and treatment response remain untested. Clinical utility also requires independent and prospective validation, a clinical operating threshold, and decision analysis.

The recurrent proteins HSPA12B, PALM2-AKAP2, STARD13, and PPM1F and the SHAP rankings generate hypotheses for tissue research. The study lacks pathway enrichment, protein-complex analysis, systematic literature comparison, and experimental corroboration. Follow-up should include targeted mass spectrometry in an independent LUAD cohort, GO/Reactome/KEGG enrichment against an appropriate protein background with multiple-testing correction, and comparison with established CPTAC-LUAD findings \citep{gillette2020}. Explainable multidimensional plasma proteomics offers a complementary research direction \citep{gushterov2026}. Independent verification should precede evaluation in settings such as benign versus malignant lesions or blood-based testing.

The matched results qualify the role of attribution in the workflow. The zero percentile cutoff made the inclusive predicate constant among all ranked proteins, while ranking still limited the candidate pool. Absolute-coefficient substitution reproduced the clauses and predictions in both analyses. The no-SHAP arm removes the predicate and changes candidate ordering, so its performance difference reflects the combined change. Signed SHAP values can still communicate sample-level contributions to the fitted logit. The study did not assess their value to human readers. The bootstrap intervals describe uncertainty conditional on the fitted out-of-fold predictions and omit variability from refitting. The reported two-of-five exact-clause recurrence is an observation from one realisation of the grouped resampling design. A single split seed per cohort provides no estimate of how recurrence varies across alternative fold assignments.

\section{Conclusion}
The analysis supports the feasibility of a three-stage explainability workflow that links patient-matched differential proteomics, sparse logistic regression with linear-model SHAP attribution, and FOL/ILP rule induction under patient-grouped evaluation. The workflow achieves high held-out discrimination for tumour versus adjacent-normal tissue within the CPTAC-LUAD cohort, but the results do not establish a clinically useful biomarker or generalisable diagnostic model. Exact symbolic rules show limited recurrence across the five outer folds. External cohort validation, repeated grouped resampling, biological pathway annotation, and experimental follow-up are necessary next steps.

Absolute L1 coefficients reproduced the SHAP-guided candidate sets, clauses, and held-out predictions in both analyses. Its paired-only comparison produced identical predictions from the full learner and a single-protein threshold. The held-out decision trace demonstrates how the rule representation exposes the threshold and evidence used for a classification. 

\end{document}